\documentclass[letterpaper, 10 pt, conference]{ieeeconf}  
\pdfoutput=1

\IEEEoverridecommandlockouts                              

\usepackage{graphics} 
\usepackage{subfigure}
\usepackage{epsfig} 
\usepackage{amsmath} 
\usepackage{amssymb}  
\usepackage[T1]{fontenc}
\usepackage{bm}
\usepackage{cite}
\usepackage{booktabs}
\usepackage{multirow}
\usepackage{diagbox}
\usepackage{threeparttable}
\usepackage{colortbl}
\usepackage{hyperref}
\hypersetup{
    colorlinks=false, 
    }
\hypersetup{hidelinks} 
\newcommand \transpose {\mathsf{T}} 

\title{\LARGE \bf
In-Hand Following of Deformable Linear Objects \\ Using Dexterous Fingers with Tactile Sensing
}

\author{Mingrui Yu\textsuperscript{1}, 
Boyuan Liang\textsuperscript{2}, 
Xiang Zhang\textsuperscript{2}, 
Xinghao Zhu\textsuperscript{2},
Lingfeng Sun\textsuperscript{2}, 
Changhao Wang\textsuperscript{2}, \\
Shiji Song\textsuperscript{1},
Xiang Li\textsuperscript{1,*}, 
and Masayoshi Tomizuka\textsuperscript{2,*}
\thanks{
\textsuperscript{1}Department of Automation, Beijing National Research Center for Information Science and Technology, Tsinghua University, China.}%
\thanks{
\textsuperscript{2}Mechanical Systems Control Lab, UC Berkeley, Berkeley, CA, USA.}%
\thanks{
\textsuperscript{*}Corresponding authors. Emails: \texttt{xiangli@tsinghua.edu.cn}, \texttt{tomizuka@berkeley.edu.cn}}%
\thanks{ This work was supported in part by the National Key Research and Development Program of China under Grant 2022YFB4701401, in part by National Natural Science Foundation of China under Grant 623B2059 and U21A20517, and in part by the Institute for Guo Qiang, Tsinghua University.
}%
}

\begin{document}

\maketitle
\pagestyle{empty}  
\thispagestyle{empty} 

\begin{abstract}
Most research on deformable linear object (DLO) manipulation assumes rigid grasping. However, beyond rigid grasping and re-grasping, in-hand following is also an essential skill that humans use to dexterously manipulate DLOs, which requires continuously changing the grasp point by in-hand sliding while holding the DLO to prevent it from falling. Achieving such a skill is very challenging for robots without using specially designed but not versatile end-effectors. 
Previous works have attempted using generic parallel grippers, but their robustness is unsatisfactory owing to the conflict between following and holding, which is hard to balance with a one-degree-of-freedom gripper. 
In this work, inspired by how humans use fingers to follow DLOs, we explore the usage of a generic dexterous hand with tactile sensing to imitate human skills and achieve robust in-hand DLO following. 
To enable the hardware system to function in the real world, we develop a framework that includes Cartesian-space arm-hand control, tactile-based in-hand 3-D DLO pose estimation, and task-specific motion design.
Experimental results demonstrate the significant superiority of our method over using parallel grippers, as well as its great robustness, generalizability, and efficiency.

\end{abstract}

\section{Introduction}

Although robotic manipulation of deformable linear objects (DLOs), such as cables and ropes, has been widely researched \cite{zhu2021challenges}, most of the existing studies assume rigid grasp of DLOs, e.g., shaping by fixedly grasping the two DLO ends \cite{yu2023generalizable,yu2022global,wang2022offline,yu2021shape,yu2023acoarse,lv2023learning} or planar rearrangement by a series of pick-and-place actions \cite{huang2023untangling,lee2022sample,jin2022robotic}. 
However, humans usually do not manipulate DLOs entirely relying on rigid grasp and re-grasp, but involving some dexterous in-hand actions, such as following the DLO towards the other end-tip by in-hand sliding, which is called \textit{DLO Following} \cite{she2021cable}. This skill is essential and efficient for many high-level tasks, such as cable routing and wrapping. However, such a task is much more challenging for a robot to handle than tasks with rigid grasp, since the robot has to achieve precise in-hand manipulation, i.e., carefully controlling the gripping motion and force to continuously change the grasp point while holding the DLO to prevent it from falling off.

\begin{figure} [tb]
  \centering 
    \includegraphics[width=8.2cm]{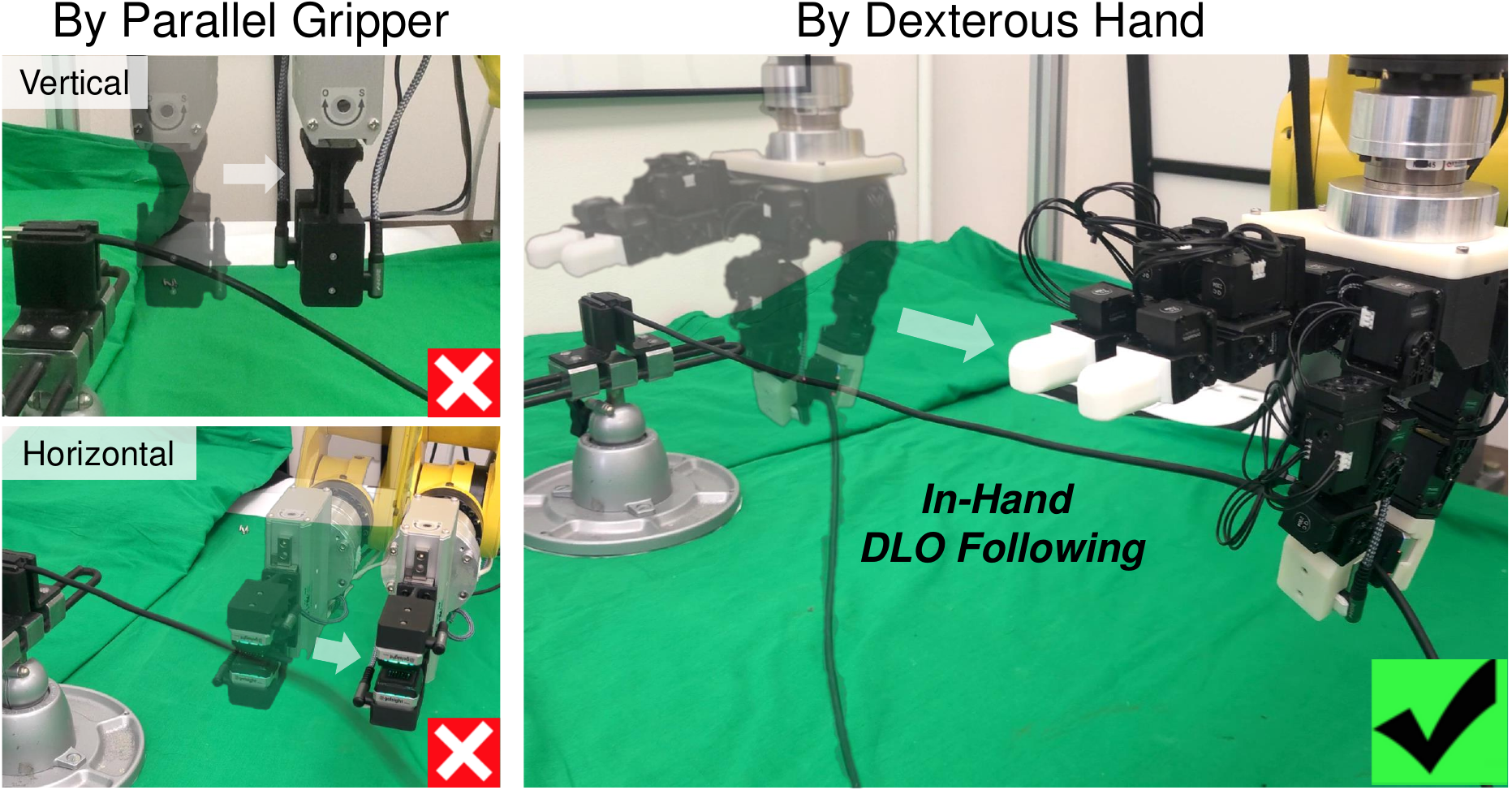} 
  \vspace{-2mm}
  \caption{Task of in-hand DLO following. The goal is to slide along the DLO towards the other end while holding the DLO. We propose a control and sensing framework to pioneeringly achieve it using a generic dexterous hand with tactile sensing, which exhibits significantly better robustness and generalizability than existing works based on parallel grippers.}
  \label{fig:fig_1}
    \vspace{-5mm}
\end{figure}

To ensure holding the DLO during following, a straight-forward approach is to design specialized end-effectors whose mechanical structures naturally guarantee holding the DLO between fingers \cite{jiang2015robotized,chapman2021alocally,wilson2023cable,suberkrub2022feel,monguzzi2024sensorless}, but then the gripper will lose its versatility and cannot manipulate other objects.
Instead, some researchers have tried using generic parallel jaw grippers for DLO following \cite{she2021cable}. However, since parallel grippers have only one degree of freedom (DoF), it is challenging to find a proper gripping force for following, as large forces will prevent in-hand sliding while small forces cannot robustly keep the DLO between fingers. 
As shown in Fig. \ref{fig:fig_1}, the parallel gripper can only be horizontally placed owing to the gravity, and frequent re-grasp by another manipulator is unavoidable (see Section \ref{sec:related_work_following} for more discussion).

Limited by the low dexterity of parallel grippers, these problems are difficult to solve from an algorithmic perspective. Instead, we try to overcome these problems from another perspective: by using dexterous hands with tactile sensors. 
Dexterous hands are also generic and versatile end-effectors but with much higher DoFs. 
Beyond the dexterous manipulation of rigid objects \cite{chen2023visual, jiang2024contact}, how to use dexterous hands to enhance DLO manipulation is still an open question. 
In this work, we study how to enable a dexterous hand with tactile sensing to achieve DLO following with much better performance, with no need for collecting offline data and learning dynamical models.
We use LEAP Hand \cite{shaw2023leaphand}, an open-sourced hand costing less than 2,000 USD. 
To the best of our knowledge, we are the first to explore DLO following based on dexterous hands in the real world. 

Although dexterous hands and tactile sensors provide the hardware foundation, additional challenges arise due to the high DoFs of the hand and noisy sensing from the tactile sensors.
To enable the hardware for deployable manipulation, we propose a framework that mainly includes three parts: 

\begin{itemize}
    \item \textbf{Control of the arm-hand system}: we design an optimization-based multi-objective inverse kinematics (IK) solver for the arm-hand system to achieve Cartesian-space control of the fingertips, and a hybrid position/force control method for the position-controlled LEAP Hand to achieve simultaneous control of finger motion and gripping forces.

    \item \textbf{Tactile estimation of in-hand DLO poses}: we propose an adaptive contact region segmentation approach and an optimization-based 3-D line fitting approach to estimate the in-hand 3-D DLO pose from two GelSight Mini tactile sensors mounted on two fingertips.

    \item \textbf{Motion design for DLO following}: based on the generic sensing and control methods, we design simple but effective robot motions to achieve DLO following with significantly better performance than the existing methods using parallel grippers.
\end{itemize}

Our control and sensing formulation can provide a generic solution to DLO following and also other tasks about in-hand DLO manipulation.
Experimental results demonstrate that our framework successfully enables the dexterous hand with tactile sensing to achieve DLO following with great robustness, generalizability, and efficiency. The supplementary materials are available on the project website \url{https://mingrui-yu.github.io/DLO_following}.

\section{Related Work} \label{sec:related_work_following}

\textbf{DLO Following}: The \textit{DLO following} task refers to sliding along the DLO towards the other end-tip. Such tasks can be divided into two categories. 
We call the first category as \textbf{\textit{shape following}}, in which both the two ends of a stiff DLO are fixed by external fixtures. The robot is required to move exactly along the DLO to scan the shape or find the other fixed end \cite{monguzzi2023tactile} without affecting the existing DLO shape.

We call the second category as \textbf{\textit{following while holding}}, in which only one end-tip of the DLO is externally fixed, and the other end sags naturally and may form arbitrary shapes. The robot needs to slide along the DLO from the fixed end towards the free end while keeping the DLO between fingers. This task is challenging if not using specially designed but not versatile end-effectors, in which the key difficulty is to balance the sliding and gripping, as large gripping forces are not allowed for in-hand sliding, but small gripping forces cannot hold the DLO.
She et al. \cite{she2021cable} was the first to try this task with generic parallel grippers. They horizontally placed the gripper because otherwise, the DLO would quickly fall off due to gravity. However, a large horizontal space will be occupied and the rotation of the end-effector along the vertical axis will be hard, which limit its applications in common table-top tasks.
In addition, they 
modeled the relationship between the gripper motion and in-hand DLO position in a data-driven manner, then dynamically controlled the gripper motion perpendicular to the DLO to keep the DLO in hand. However, such a model is affected by many factors, such as the DLO's plastic deformation, shapes, stiffness, and frictions, so it is theoretically difficult to generalize well to arbitrary DLO. 
Considering the inevitable modeling error and small control error tolerance (i.e., small tactile-sensing area),
the DLO is easy to fall off without frequent re-grasp by another manipulator.
Reinforcement learning (RL) has been also adopted, but only validated in simulation \cite{pecyna2022visual} or generalized poorly on different types of DLOs \cite{hellman2017functional}.

\textbf{Tactile-Based DLO Sensing}: Vision-based DLO global state estimation has been widely studied \cite{lv2023learning,zhaole2023robust}.
However, owing to occlusions and depth errors, vision can not precisely estimate the in-hand DLO states, which is required for controlling the interaction between the end-effector and DLO.
In contrast, tactile sensors can directly acquire the in-hand contact states. 
In \cite{she2021cable, wilson2023cable}, the Principal Component Analysis (PCA) method is employed to estimate the DLO pose on the 2-D surface of a GelSight \cite{yuan2017gelsight} for DLO following.
The 2-D in-hand DLO states are also estimated by fitting or learning-based methods for DLO insertion \cite{pirozzi2018tactile} and needle-threading \cite{yu2023precise}.
Beyond 2-D states, in-hand 3-D DLO poses are indispensable for DLO following in 3-D spaces.
In \cite{monguzzi2023tactile}, complicated procedures and rules are manually designed to roughly estimate the 3-D states, but the implementation is cumbersome and active robot motion is required during estimation. 
In contrast, we propose an elegant optimization-based approach to estimate the in-hand 3-D DLO pose at any moment from two tactile sensors.

\textbf{Dexterous Manipulation of Deformable Objects}: 
Dexterous hands are more powerful generic end-effectors than parallel grippers, but 
few studies explore their potential in deformable object manipulation.
A local deformation control method for 3-D soft objects by a dexterous hand was proposed in \cite{ficuciello2018fem}. 
A twisting skill for ropes was implemented in \cite{takizawa2016implementation} based on specially designed fingers.
A very recent work \cite{zhaole2023dexdlo} learned RL-based policies of a fixed-base Shadow Hand for in-hand DLO manipulation; however, it is evaluated purely in simulation, and sim-to-real transfer is challenging.  

\section{Methodology}

\begin{figure} [tb]
  \centering 
    \includegraphics[width=8.5cm]{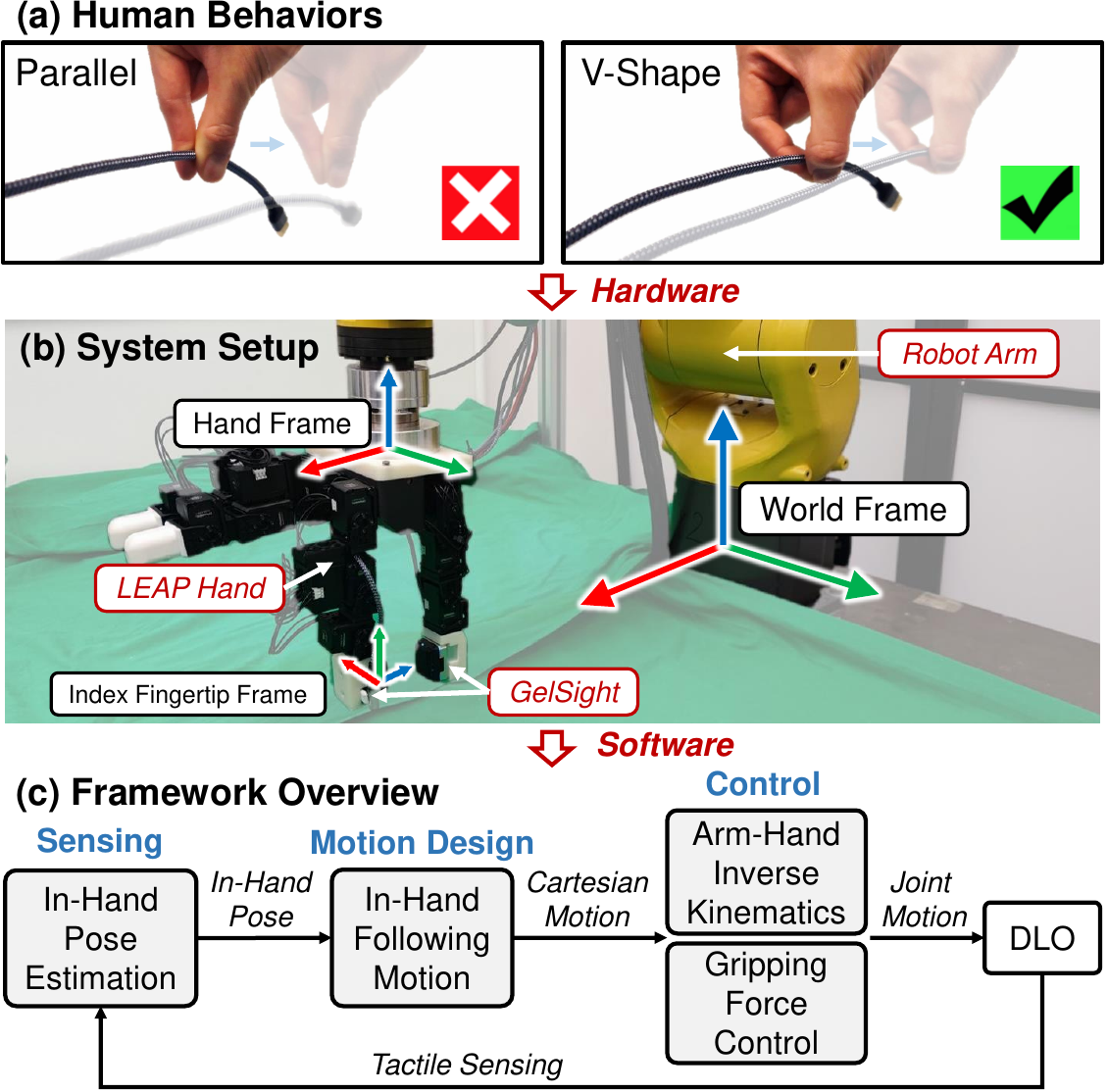} 
  \vspace{-1mm}
  \caption{Overview of this work's inspiration, hardware setup, and algorithm framework. 
  (a) How humans use fingers to follow DLOs. Humans usually form the two fingers as a V-shape during DLO following; otherwise, the DLO will easily fall off owing to gravity, even if manipulated by humans. 
  (b) The hardware setup and the defined frames. 
  The thumb-tip frame is symmetrical to the index-fingertip frame. 
  For each frame, the red, green, and blue arrow refers to the X, Y, and Z axis, respectively. 
  (c) Our algorithm framework to enable DLO following by the dexterous hand, which includes generic control, sensing, and task-specific motion design.}
  \label{fig:framework_structure}
    \vspace{-3mm}
\end{figure}

In this work, the key idea for enhancing DLO following by a dexterous hand is inspired by how humans use fingers to follow DLOs in daily life. As shown in Fig. \ref{fig:framework_structure}(a), humans usually form the index and thumb fingers as a V-shape, which ensures holding the DLO without applying large contact forces to the DLO. This work aims to imitate such human skills in DLO following by a generic dexterous hand with tactile sensing.
Fig. \ref{fig:framework_structure}(b) shows the hardware setup and defined frames. It contains an anthropomorphic LEAP Hand installed on a robot arm and two GelSight Mini 
tactile sensors mounted on the index and thumb fingertips. In this work, we explore the usage of the index and thumb fingers and keep the other two fingers fixed. We denote the world frame and hand frame as $\mathcal{W}$ and  $\mathcal{H}$, respectively.

To enable human-like in-hand following of DLOs by the above robot system, we propose an algorithm framework involving generic arm-hand control, tactile sensing, and task-specific motion design, as overviewed in Fig. \ref{fig:framework_structure}(c). 
First, the in-hand 3-D DLO pose is estimated from the original tactile sensing; next, the desired Cartesian-space motion for DLO following is determined according to the in-hand state; then, the proposed control method maps the Cartesian-space motion to the robot joint space and drive the robot to interact with the DLO. These three parts are introduced below.

\subsection{Cartesian-Space Control of the Arm-Hand System}

As for the low-level control module, two main objectives need to be achieved: controlling the Cartesian-space motion of fingertips and the gripping forces between fingertips. For the former, we design an optimization-based multi-objective IK solver; for the latter, we design a hybrid position/force control method for the position-controlled LEAP Hand.

\subsubsection{Optimization-Based IK Solver} \label{sec:ik}

Mapping the desired Cartesian-space motion of fingertips to the joint space is non-trivial. Although this manipulator with 14 DoFs seems to be a redundant system, it often exhibits over-constrained characteristics. 
First, the 6-D finger pose in the hand frame is under-actuated since each finger has 4 DoFs. Many works simplify the contact between a fingertip and the object to a point so ignore the fingertip orientation. However, in our problem, the fingertip orientation highly affects the quality of grasping and tactile sensing for DLOs.
Second, the workspace of fingertips is highly constrained by the hand joint limits and self-collision, so the popular inverse Jacobian methods for IK solving easily get stuck at the joint bounds \cite{beeson2015trac}.
Third, multiple Cartesian-space objectives need to be considered for DLO following, and there may be conflicts between them, 
e.g., both the fingertip poses in $\mathcal{W}$ and $\mathcal{H}$ as well as the hand pose in $\mathcal{W}$.
Null-space-based methods (e.g. \cite{qiu2021precision}) are not suitable since it is difficult to explicitly distinguish higher-level and lower-level objectives.
Consequently, in this work, we formulate the IK solving as a constrained optimization problem, in which all objectives are simultaneously combined by weighting coefficients.

The potential Cartesian-space objectives include: the poses of the two fingertips in $\mathcal{W}$ (i.e., $^{w}{\bm P_{f, i}}$), the poses of the two fingertips in $\mathcal{H}$ (i.e., $^{h}{\bm P_{f, i}}$), the relative position between the two fingertips in $\mathcal{H}$ (i.e., $^{h}\Delta \bm p_f = {^{h}\bm p_{f, 1}} - {^{h}\bm p_{f, 2}} $), and the hand pose in $\mathcal{W}$ (i.e., $^{w}{\bm P_h}$). For different motions, different objectives can be included with different weights. 
The optimization-based IK solving is formulated as
\begin{equation} \label{eq:optimization_ik}
\begin{aligned}
      \min_{\bm q} \,  
       & \mathcal{C} = \frac{1}{2} \sum_{i=1}^{2} {^{w}\tilde{\bm P}_{f, i}^\transpose} \bm W_{fw, i} {^{w}\tilde{\bm P}_{f, i}}
       + \frac{1}{2} \sum_{i=1}^{2} {^{h}\tilde{\bm P}_{f, i}^\transpose} \bm W_{fh, i} {^{h}\tilde{\bm P}_{f, i}}
       \\ 
       & + \frac{1}{2} {^{h}\tilde{\Delta \bm p}_f^\transpose} \bm W_{rfh} {^{h}\tilde{\Delta \bm p}_f}
     + \frac{1}{2} {^{w}\tilde{\bm P}_{h}^\transpose} \bm W_{hw} {^{w}\tilde{\bm P}_{h}}
    \\
    \text{s.t.} 
    & \quad \bm q_{\rm lb} \leq \bm q \leq \bm q_{\rm ub} 
\end{aligned}
\end{equation}
where $\tilde{(\cdot)}$ refers to the error between the desired pose and the current pose calculated by the joint position $\bm q$ and  robot forward kinematics. Each objective and each dimension of the objective can be weighted by the weighting matrices $\bm W_{fw}, \bm W_{fh}, \bm W_{rfh}$ and $\bm W_{hw}$. The joint positions are constrained by the lower bound $\bm q_{\rm lb}$ and upper bound $\bm q_{\rm ub}$.
We use the Sequential Least Squares Programming (SLSQP) \cite{kraft1988software} algorithm to solve this optimization problem.

The advantage of this approach is that it can specify all potential objectives in a unified formulation. Additionally, it allows addressing the conflicts between different objectives in a soft manner, which can be easily adjusted by changing the weighting matrices (see Section \ref{sec:exp_ik} for the specific objectives and weights we used for DLO following).

\subsubsection{Hybrid Position/Force Control for LEAP Hand}

Controlling the gripping force between the two fingertips is essential for grasping and sliding along the DLO.
The default controller of the open-sourced LEAP Hand is a joint position controller, and the maximum communication frequency is less than 90 Hz. Therefore, conventional torque control approaches cannot be directly applied, such as impedance control \cite{wimbock2007impedance,yoshikawa2010multifingered}. Thus, we design a gripping force control method based on the hand's default position controller.

The hand's default position controller adopts the current-based PID position control mode of the Dynamixel motors. 
The joint torques $\bm \tau$ are approximately related to the desired joint positions $\bm q_d$ and current joint positions $\bm q$ as
$\bm\tau = k_p \bm (\bm {q}_d - \bm{q}) - k_v \dot{\bm q}$ when the integral gain is set to zero. Since stable gripping is quasi-static, we assume $k_v \dot{\bm q} = \bm 0$, and we assume the external force acting on the joints of a finger is equal to the output joint torques as
$\bm \tau = \bm J(\bm q)^\transpose \bm F_{\rm ext}$.
Then, we can enable a fingertip to exert force $\bm F_d$ to the object by assigning the desired joint position $\bm q_d$ of the finger as
\begin{equation} \label{eq:force_control}
    \bm q_d = \bm q + k_p^{-1} \bm J(\bm q)^\transpose \bm F_{d}
\end{equation}

\begin{figure*} [tb]
  \centering 
\includegraphics[width=0.90\textwidth]{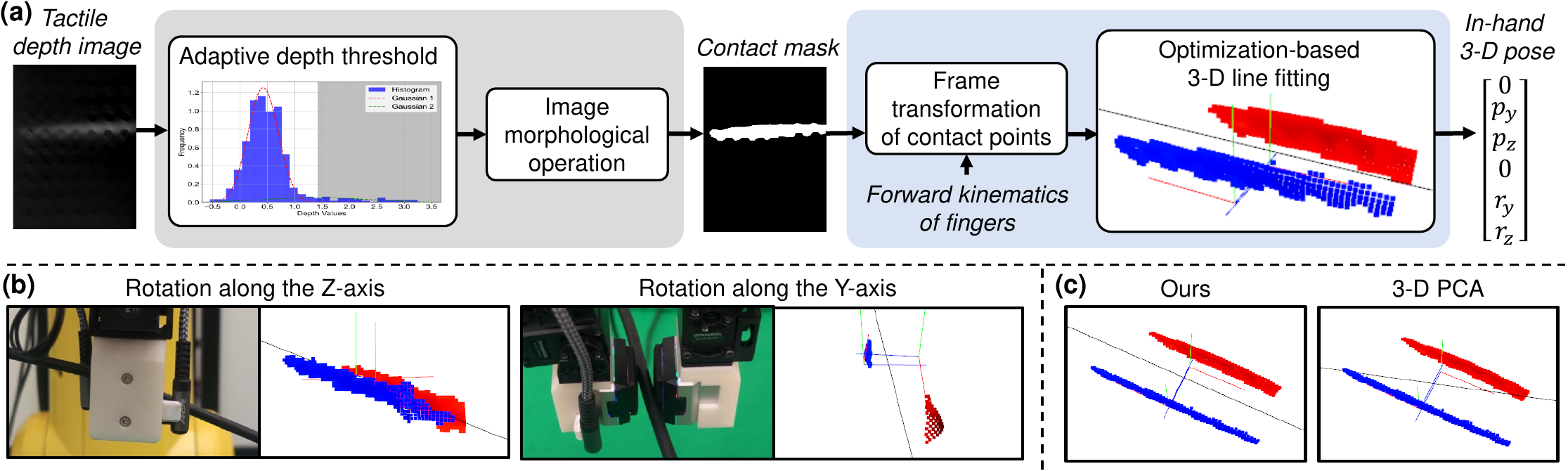} 
  \vspace{-3mm}
  \caption{Tactile sensing of the in-hand DLO pose. (a) Estimation pipeline. (b) Estimation results of the in-hand DLO that rotates along the Z and Y axes of the index fingertip frame. (c) Comparison between our optimization-based 3-D line fitting method and the 3-D PCA method. The blue and red points represent the contact points on the index and thumb GelSight, respectively; the black line represents the estimated DLO axis.}
  \label{fig:sensing_pipeline}
    \vspace{-3mm}
\end{figure*}

Moreover, we need to simultaneously control the orientation of the fingertips during DLO following. Thus, to form a hybrid position/force controller for gripping, we first use the proposed IK solver to calculate the target joint position $\bm q_t$, then add the force term. Replacing $\bm q$ in (\ref{eq:force_control}) by $\bm q_t$, we obtain the final joint position command sent to the hand.

\subsection{Tactile Sensing of the In-Hand DLO State}

In-hand DLO poses are required by DLO following and many other tasks. 
The pipeline of our tactile sensing approach is illustrated in Fig. \ref{fig:sensing_pipeline}(a).
Compared with 2-D poses on one tactile sensor, estimating 3-D poses from two sensors is necessary since 1) the 3-D in-hand pose can guide the robot to adjust its 3-D motion for DLO following, and 2) the contact region on only one sensor can be too small for reliable pose estimation (e.g., Fig. \ref{fig:sensing_pipeline}(b) right).

\subsubsection{Contact Region Segmentation}

A straightforward method to segment contact regions from depth images of the GelSight is using a constant depth threshold \cite{she2021cable}. However, we find that the depth values of non-contact regions are highly affected by contact positions and forces. Thus, we adopt a flexible depth threshold. 
We apply the Expectation-Maximization (EM) algorithm to approximate the distribution of all depth values by a two-component Gaussian mixture model (GMM). We assume the Gaussian component $\mathcal{N}(\mu, \sigma)$ with a higher weight represents the non-contact region. Then, the depth threshold for contact segmentation is specified as $D_{\rm thres} = \mu + 3\sigma$. 
We also apply image morphological operations to further process the contact masks.
In addition, when the two fingers form a V-shape, the bottoms of the two sensors contact each other, so we manually remove the bottom region from the segmented contact masks.

\subsubsection{In-Hand DLO Pose Estimation} \label{sec:inhand_pose_estimation}

After extracting the contact regions, we use the forward kinematics of fingers to transform all contact points from two sensors to the same frame.
This set of 3-D contact points represents the contact DLO surface. 
We define the in-hand DLO pose as a 3-D line in the index fingertip frame, whose distances to all contact points are as close to the DLO radius $r$ as possible.
A 3-D line can be parameterized as $\bm \xi + t \bm \psi$, where $\bm \xi \in \mathbb{R}^3$ is a point on the line and $\bm \psi \in \mathbb{R}^3$ is the line direction. The distance $d_i$ between a point $\bm o_i$ and the line is calculated as $d_i  = \frac{ \| (\bm o_i - \bm \xi) \times \bm \psi \|}{\| \bm \psi \|}$.
Basically, we can formulate an optimization problem whose variable is $(\bm \xi, \bm \psi, r)$ and cost is $\sum_{i=1}^{n} (d_i - r)^2$ where $n$ is the number of the contact points. However, due to the inaccuracy of the forward kinematics of fingers, we allow the optimization to slightly adjust the relative position $\Delta \bm p \in \mathbb{R}^3$ between the two tactile frames.
Then, the optimization variable is defined as 
$\bm \Theta = (\bm \xi, \bm \psi, r, \Delta \bm p)$.
We also add regularization terms for the variables.
The final optimization problem is formulated as 
\begin{equation} \label{eq:optimization_dlo}
\begin{aligned}
      \min_{\bm \Theta}  \quad  
       & \frac{1}{2} \sum_{i=1}^{n} (d_i(\bm \Theta) - r)^2 
       + \frac{1}{2} \bm \Theta^\transpose  \bm W_{r} \bm \Theta
    \\
    \text{s.t.} 
    \quad & \| \bm \psi \|_2^2 = 1
    , \quad \xi_x = \sum_{i=1}^{n} o_{i, x}
\end{aligned}
\end{equation}
where $\bm W_r$ is the weight of regularization. 
Note that a line has four DoFs although its parameter is 6-dimensional, so we constrain the norm of the direction $\bm \psi$ to be equal to $1$ and the x-axis position of $\bm \xi$ (i.e., $\xi_x$) to be equal to the average x-axis position of all contact points (i.e., $\sum_{i=1}^{n} o_{i, x}$). 
We use the SLSQP algorithm to solve this problem. The 4-dimensional pose can then be calculated based on $\bm \xi$ and $\bm \psi$.

Previous works usually use the 2-D PCA method to estimate the main axis on a 2-D surface. However, the 3-D PCA cannot be adopted for our problem because of the spatial gap between the contact points from the index and thumb fingertips. As shown in Fig. \ref{fig:sensing_pipeline}(c), the 3-D PCA will find a line that minimizes the absolute distances to all points instead of minimizing the differences between the distances.

\subsection{Motion Design for DLO Following}

Based on our arm-hand control and in-hand state estimation framework, we design the Cartesian-space robot motions to imitate human's finger actions for DLO following.

\begin{figure} [tb]
  \centering 
    \includegraphics[width=7cm]{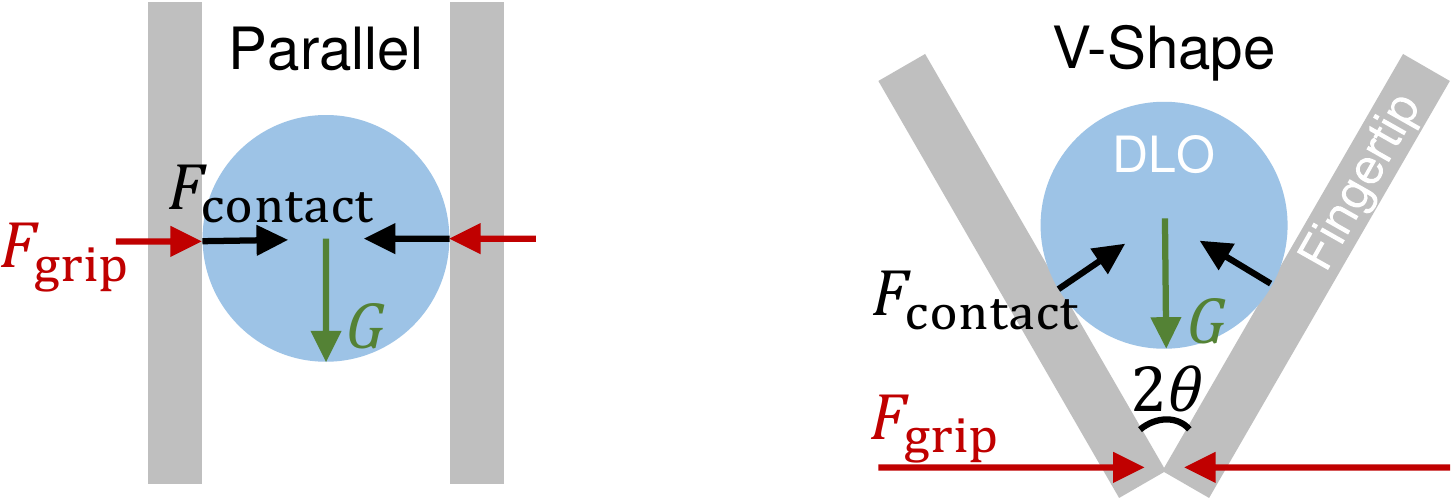} 
  \vspace{-3mm}
  \caption{Superiority of V-shape fingers over parallel fingers. In the parallel setting, the contact force $F_{\rm contact}$ can only equal the gripping force $F_{\rm grip}$. In contrast, these two can be separately controlled by V-shape fingers.
  }
  \label{fig:force_analysis}
\end{figure}

As illustrated in Fig. \ref{fig:force_analysis}, it is difficult for parallel grippers to find a proper gripping force to balance following and holding. In contrast, by utilizing the more DoFs of dexterous fingers, the two objectives can be decoupled by forming a ``V'' shape: for holding, it can apply a large gripping force to keep the bottom of the two fingertips tightly close; and for following, it can adjust the angle of the V shape (i.e., gripping angle $\theta$) to apply a moderate contact force. 

The gripping angle $\theta$ for a specific DLO is initially determined during grasping: the fingers first grasp the DLO in parallel; then, the gripping angle is increased until any tactile sensor loses contact with the DLO. 
During the following, the gripping angle will be increased if the shear forces on the tactile sensors are larger than a threshold and be reduced if the in-hand contact does not satisfy the requirements of reliable sensing.
We restrict the gripping angle between $15^{\circ}$ and $50^{\circ}$ and set the gripping $F_{\rm grip}$ to a constant value.

\begin{figure} [tb]
  \centering 
    \includegraphics[width=8.0cm]{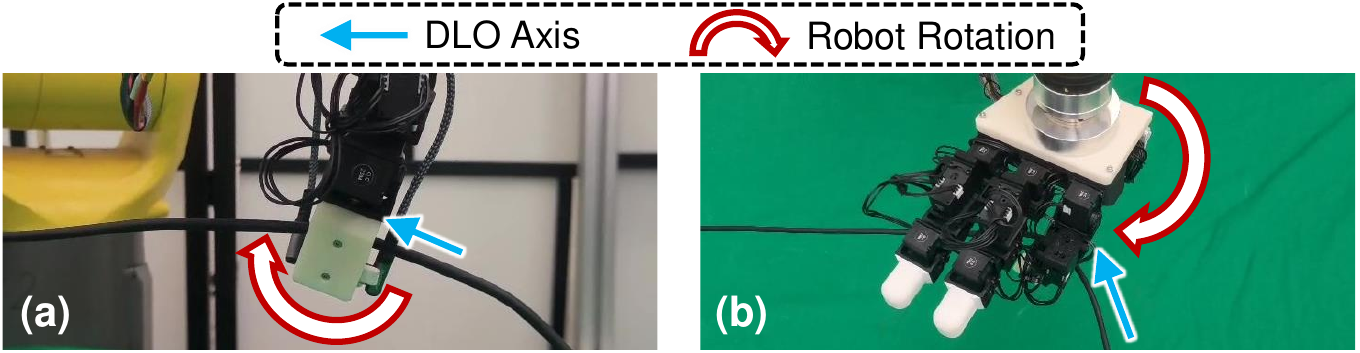} 
  \vspace{-2mm}
  \caption{Orientation adjustment to better match the in-hand DLO pose. (a) Along the X-axis of the hand frame. (b) Along the Z-axis of the hand frame.}
  \label{fig:rotation_adjust}
    \vspace{-3mm}
\end{figure}

Moreover, we find it is necessary to adjust the orientations of the fingertips to match the in-hand 3-D DLO pose during the following; otherwise, the tension or bending forces along the DLO will easily make the two fingertips misaligned. 
Thus, we adopt a proportional (P) controller to control the rotation of the fingertips along the X and Z axes of $\mathcal{H}$ to better match the estimated in-hand DLO pose (see Fig. \ref{fig:rotation_adjust}).

It is worth emphasizing that, compared with dynamically controlling the horizontal gripper motion to keep the DLO in hand when using parallel grippers \cite{she2021cable}, by using the dexterous hand, we can directly guarantee the holding of the DLO by the fingers' geometrical configuration, which is significantly more robust. 
In addition, the motion of the arm end-effector can be separately designed for high-level task requirements, such as for cable routing.

\section{Experimental Results}

We use a FANUC LR Mate 200iD manipulator. 
For convenience, we define a \textit{reference hand configuration} for gripping, in which the index and thumb fingertip contacts each other approximately in parallel. 
The SLSQP algorithm is implemented by the \texttt{scipy.optimize.minimize} in Python.
More details can be found on the \href{https://mingrui-yu.github.io/DLO_following}{project website}.

\subsection{IK Solver for the Arm-Hand System} \label{sec:exp_ik}

\begin{figure} [tb]
  \centering 
    \includegraphics[width=8.6cm]{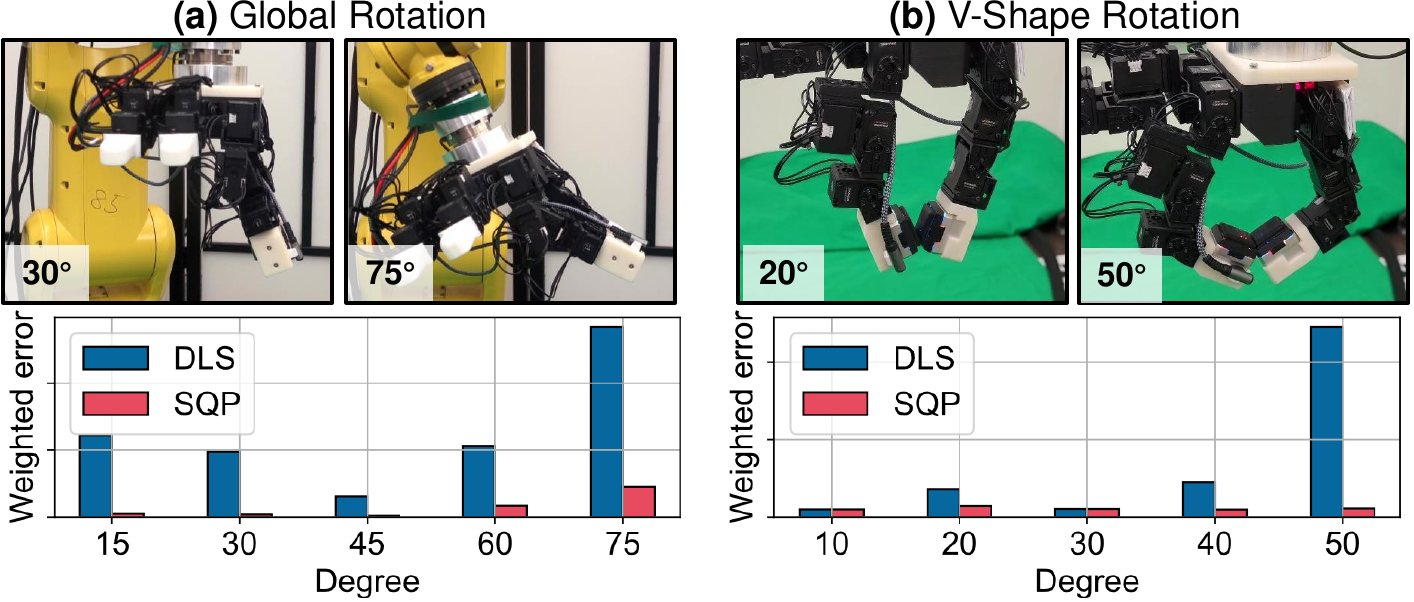} 
  \vspace{-7mm}
  \caption{Validation of the IK solver. (a) Rotating the fingertips w.r.t the world frame. (b) Rotating the fingertips to form different gripping angles. The weighted error is $\sqrt{2\mathcal{C}}$ of (\ref{eq:optimization_ik}), which does not have a specific unit.}
  \label{fig:exp_IK}
    \vspace{-2mm}
\end{figure}

Two tasks related to DLO following are considered to validate the proposed IK solver. 
The first is \textit{global rotation}, in which the target is to (only) change the orientation of the two closed fingertips along the X-axis of the world frame (see Fig. \ref{fig:exp_IK}(a)). Such motion will be used in the $\textit{shape following}$ task (see Section \ref{sec:shape_following}).
The objectives of (\ref{eq:optimization_ik}) are specified as 1) the desired 6-D pose of the index fingertip in $\mathcal{W}$, with the weight $\bm W_{fw, 1} = \text{diag}(100, 100, 100, 1, 1, 1)$, 2) the relative position between the two fingertips in $\mathcal{H}$ to be zero, with $\bm W_{rfh} = \text{diag}(10,10,10)$, 3) the orientations of two fingertips along the Y and Z axis of $\mathcal{H}$ to be close to those of the reference hand configuration, with $\bm W_{fh, i}=\text{diag}(0,0,0,0,0.1,0.1)$, and 4) the rotation of the hand along the X and Y axis of $\mathcal{W}$ to be close to zero, with $\bm W_{hw} = \text{diag}(0, 0, 0, 0.05, 0.05, 0.05)$. Fig. \ref{fig:exp_IK}(a) shows that the robot will first rotate the fingers to reach the desired orientation to fully utilize the hand dexterity; when the fingers reach their joint limits, the robot will start to rotate the arm end-effector. 

The second task is \textit{V-shape rotation}, where the target is to change the gripping angle while maintaining the pose of the center of two fingertips in $\mathcal{W}$ (see Fig. \ref{fig:exp_IK}(b)). This is the essential motion for the \textit{following while holding} task. 
The objectives of (\ref{eq:optimization_ik}) are specified as 1) the desired fingertip poses in $\mathcal{H}$, with $\bm W_{fh, i}=\text{diag}(1,0,0,0.1,0.1,0.01)$, 2) the relative position between the two fingertips to be zero, with $\bm W_{rfh} = \text{diag}(100,100,100)$, and 3) the fixed position of the index fingertip in $\mathcal{W}$, with $\bm W_{fw, 1} = \text{diag}(100, 100, 100, 0, 0, 0)$, and 4) the fixed orientation of the hand in $\mathcal{W}$, with $\bm W_{hw} = \text{diag}(0, 0, 0, 10, 10, 10)$. Fig. \ref{fig:exp_IK}(b) shows that the fingers can form a gripping angle from $0^{\circ}$ to $50^{\circ}$.

We compare the proposed IK solver based on sequential quadratic programming (SQP) with the classical weighted damped least square (DLS) method \cite{buss2004introduction}. The results in Fig. \ref{fig:exp_IK} indicate that the DLS performs poorly in such constrained multi-objective IK problems, while our method obtains precise and stable results. 

\begin{figure} [tb]
  \centering 
    \includegraphics[width=8.6cm]{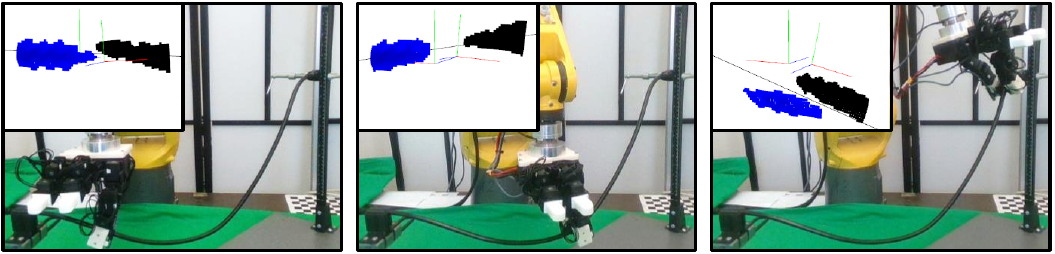} 
  \vspace{-6mm}
  \caption{Shape following task on a two-end-fixed DLO to validate both the proposed IK solver and in-hand DLO pose estimation approach. The tactile-based estimation results are visualized in the upper left figures.}
  \label{fig:exp_shape_following}
    \vspace{-4mm}
\end{figure}

\subsection{Shape Following Task} \label{sec:shape_following}

We validate both the proposed IK solver and the in-hand DLO pose estimation method on the $\textit{shape following}$ task described in Section \ref{sec:related_work_following} (see Fig. \ref{fig:exp_shape_following}). Since the goal is to move exactly along a fixed DLO, accurate in-hand pose estimation is essential to determine where the robot should go, and an accurate IK solver is essential to ensure reaching the desired pose. We test on a 3-D curved thick power cable and a thin USB cable for five times, respectively. The robot successfully reaches the other end-tip of the DLO in all tests.

\subsection{Following While Holding Task}

Unless otherwise stated, the tests are performed on a typical USB cable with the desired following direction being the positive Y-axis of $\mathcal{W}$ and a following speed of 0.025 m/s.

\subsubsection{Comparison with Parallel Grippers}

\begin{table}
\centering
\caption{Comparison between dexterous hands and parallel grippers.}
\vspace{-2mm}
\label{tab:exp_comparison_with_parallel}
\begin{threeparttable}[b]
\begin{tabular}{@{\hspace{0.2em}}c@{\hspace{0.8em}}c|ccc@{\hspace{0.2em}}} 
\toprule
\begin{tabular}[c]{@{}c@{}}DLO\\shape\end{tabular} & Metrics & \begin{tabular}[c]{@{}c@{}}Parallel\\(vertical)\end{tabular} & \begin{tabular}[c]{@{}c@{}}Parallel\\(horizontal)\end{tabular} & \begin{tabular}[c]{@{}c@{}}Dexterous\\hand\end{tabular} \\ 
\hline
\multirow{2}{*}{Straight} & Follow. length (m)\tnote{a}~$\uparrow$ & 0.08 & 0.98 & \textbf{1.0} \\
 & Mov. range (m)\tnote{b}~$\downarrow$ & 0 & 0.21 & 0 \\
\multirow{2}{*}{Curved} & Follow. length\tnote{a} (m)~$\uparrow$ & 0.07 & 0.27 & \textbf{1.0} \\
 & Mov. range (m)\tnote{b}~$\downarrow$ & 0 & 0.11 & 0 \\
\bottomrule
\end{tabular}
\begin{tablenotes}
     \item[a] Ave. following length along the DLO. The max. length is 1.0 m.
     \item[b] Ave. required moving range perpendicular to the following direction.
\end{tablenotes}
\end{threeparttable}
\vspace{-3mm}
\end{table}

We first compare the proposed DLO following method using a dexterous hand with the conventional approaches using parallel grippers. As shown in Fig. \ref{fig:fig_1}, two settings using parallel grippers are tested: in the first setting, the gripper is vertically placed \cite{hellman2017functional} and moves in an open-loop manner; in the second setting, the gripper is horizontally placed and reactively moves according to the in-hand DLO pose using the same LQR-based approach in \cite{she2021cable}. 
We test the three approaches on a cable with a straight shape and a curved shape, respectively. Each approach and each shape is tested five times, and the results are summarized in Table \ref{tab:exp_comparison_with_parallel}. The vertical parallel gripper loses contact with the DLO quickly. Although the horizontal parallel gripper using a reactive controller follows a straight DLO well, it requires large movements perpendicular to the desired following direction and occupies a large horizontal space, which is impractical for table-top applications. Additionally, it cannot handle curved DLOs well. In contrast, our approach based on a dexterous hand shows robustness and practicability, as the hand reaches the maximum following length in all tests with no need for deviating from the desired following direction.

\subsubsection{Generalization to Different DLOs}

\begin{figure} [tb]
  \centering 
\includegraphics[width=\linewidth]{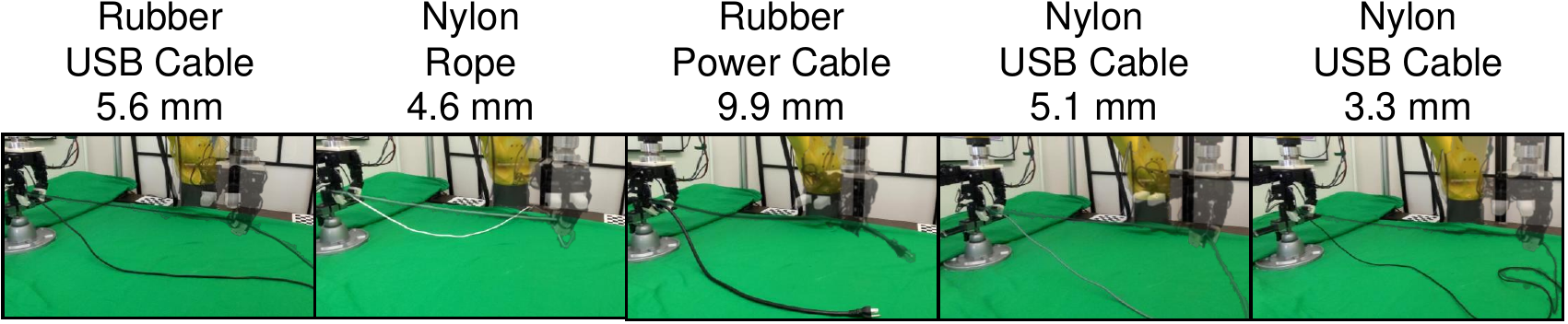} 
  \vspace{-6mm}
  \caption{Experiments on five different DLOs. The materials and diameters of the DLOs are listed above the figures. Each DLO is tested five times, and the DLO is randomly placed on the table at the beginning of each test.
  }
  \label{fig:diff_dlos}
\end{figure}

We validate the proposed approach on five different DLOs with different materials and diameters, as shown in Fig. \ref{fig:diff_dlos}. We test five times for each DLO, and we randomly place the DLO on the table in different curved shapes at the beginning of each test. Our approach succeeds in all tests, indicating great generalizability to different DLOs and shapes.

\subsubsection{Generalization to Different Following Speed}

We test the proposed method with the following speeds of 0.025, 0.05, and 0.10 m/s for five times, respectively. All tests are successful, demonstrating the robustness of our method even at very high following speeds.

\subsubsection{Comparison Between Whether Using Tactile Feedback}

\begin{figure} [tb]
  \centering 
    \includegraphics[width=8.6cm]{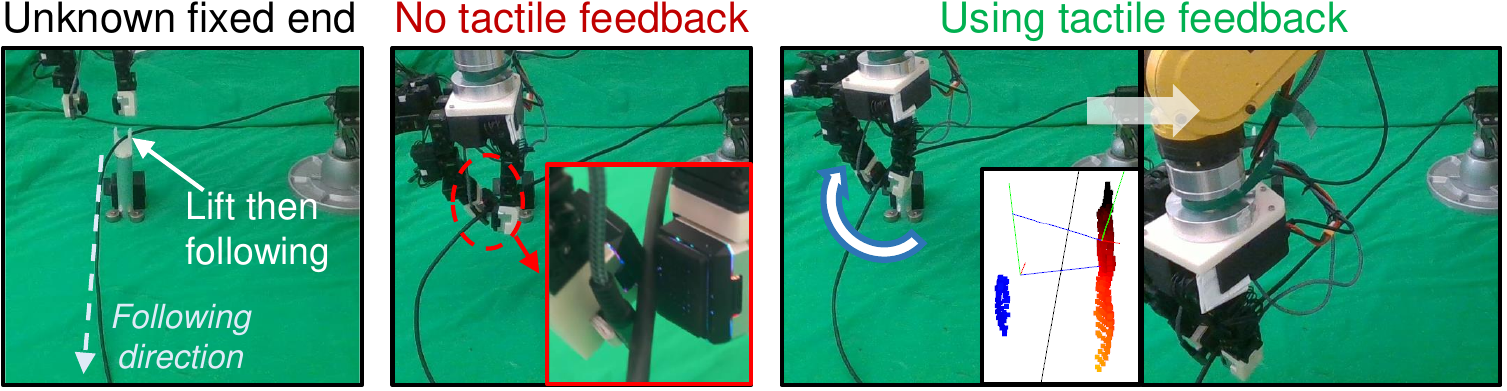} 
  \vspace{-6mm}
  \caption{Comparison between whether using tactile feedback. The task is to lift the DLO and then follow it along the desired following direction, during which the position of the fixed end is unknown.}
  \label{fig:exp_compare_tactile_feedback}
    \vspace{-3mm}
\end{figure}

We demonstrate the necessity of in-hand DLO pose estimation during DLO following in the task shown in Fig. \ref{fig:exp_compare_tactile_feedback}, whose goal is to lift the DLO and then follow it along the desired following direction, during which the position of the fixed end is unknown. In an open-loop manner without tactile feedback, the orientations of the fingers are fixed. Since the stiffness of the hand motor's position controller is limited, the two fingers become misaligned quickly owing to the oblique DLO tension force and fail to continue following.
In contrast, when using the estimated in-hand 3-D DLO pose as feedback, the fingers can reactively rotate to better match the direction of the in-hand DLO to reduce the impact of in-hand forces, and all five tests finally succeed.

\subsubsection{Switch Between Grasping and Following}

\begin{figure*} [tb]
  \centering 
\includegraphics[width=\textwidth]{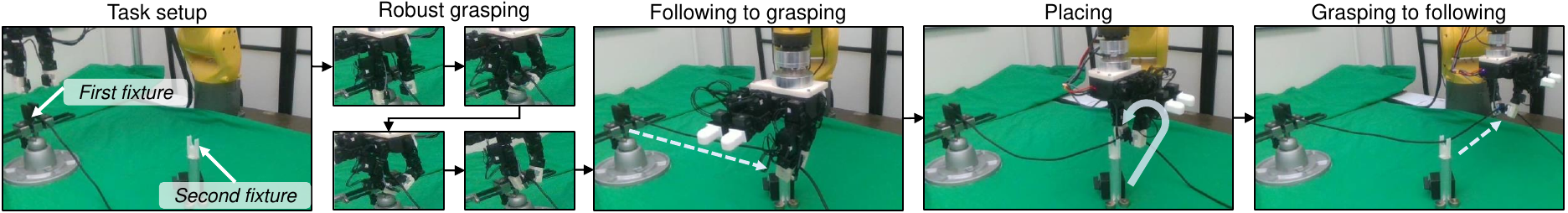} 
  \vspace{-6mm}
  \caption{A simplified high-level cable routing task that involves switching between rigid grasping and following. The robot is required to grasp the DLO beside the first fixture, follow it to a desired position, re-grasp it to place it on the second fixture, and re-follow it along another direction.}
  \label{fig:exp_final_demo}
    \vspace{-3mm}
\end{figure*}

We conduct a simplified cable routing task (Fig. \ref{fig:exp_final_demo}) to demonstrate the ability of our approach to switch between rigid grasping and following in high-level tasks. The task includes the following procedures: 1) for the initial grasping, we implement a robust grasping strategy similar to human behaviors, which can handle some initial position uncertainty;
2) the robot follows the DLO to a pre-defined position;
3) the robot switches from following to rigid grasping by changing the gripping angle to zero;
4) the robot places the DLO on the second fixture;
and 5) the robot switches from grasping to following and follows the DLO along another desired direction.
All waypoints are manually defined. We test for five times, and all succeed. 
This experiment shows the application prospects of our approach on more complicated high-level tasks.

\subsubsection{Failure cases} \label{sec:fail_cases}

Our designed motions for DLO following cannot well handle some scenarios. First, when the direction of the free part of a heavy DLO is opposite to the desired following direction, the DLO may force the fingers to open and then fall off.
Second, when an upward disturbance is exerted on the DLO, the DLO may move away from the tactile surface, causing in-hand sensing to be lost.

\section{Conclusion and Discussion}

\textbf{Conclusion}: 
This work explores the usage of a dexterous hand with tactile sensing to enhance the in-hand DLO following, which is challenging for existing approaches based on parallel grippers. 
To enable the hardware system containing arm, hand, and tactile sensors to reliably achieve this task in the real world, we propose an algorithm framework including the arm-hand control, tactile sensing, and human-like motion design.
The experimental results demonstrate the significant superiority of our method over using parallel grippers, the robustness to different DLOs and speeds, as well as the potential to be applied in high-level practical tasks.

\textbf{Discussion}:
It is worth sharing the practical problems we met: 
1) the forward kinematics of the LEAP Hand is not very accurate, as there exist slight passive movements that are not along the joint axis; 
2) the output torque of the hand motor is limited, so it cannot provide large stiffness for very rigid grasp;
and 3) the delay of receiving depth images from GelSight Mini on our computer is larger than 70 ms, which is large for handling dynamic situations.
Affected by many practical factors, the current approach cannot well handle the situations in Section \ref{sec:fail_cases}, and it is hard to precisely control the in-hand contact during the following owing to the dynamic in-hand DLO motion.
One promising direction for future work is to study how to use more fingers and the palm, which is more close to human's dexterous behaviors.








{
\bibliographystyle{IEEEtran}
\bibliography{ref}
}


\end{document}